\documentclass[lettersize,journal]{IEEEtran}
\usepackage{amsmath,amsfonts}
\usepackage{algorithmic}
\usepackage{algorithm}
\usepackage{array}
\usepackage[caption=false,font=normalsize,labelfont=sf,textfont=sf]{subfig}
\usepackage{textcomp}
\usepackage{stfloats}
\usepackage{url}
\usepackage{verbatim}
\usepackage{graphicx}
\usepackage{bm}
\usepackage{cite}
\usepackage{booktabs}
\usepackage{multirow}
\usepackage{color}
\usepackage{xcolor}
\usepackage{colortbl}

\newcommand{\added}[1]{\textcolor{black}{#1}}

\begin{document}

\title{Dual-Granularity Semantic Prompting for Language Guidance Infrared Small Target Detection}

\author{
Xuelin Qian,~\IEEEmembership{Member,~IEEE}, Zixuan Wang, Haoran Sun, Jiaming Lu, Zhongling Huang, Dingwen Zhang,~\IEEEmembership{Member,~IEEE}, and Junwei Han,~\IEEEmembership{Fellow,~IEEE}
% \thanks{This work was supported in part by the National Natural Science Foundation of China under Grant U2541228, 62406252, 62322605, and by the Guangdong Basic and Applied Basic Research Foundation (2025A1515010158). \textit{Corresponding author: Xuelin Qian}.}
% \thanks{Xuelin Qian is with the Brain and Artificial Intelligence Laboratory, School of Automation, Northwestern Polytechnical University, Xi'an 710021, China, and also with the Shenzhen Research Institute of Northwestern Polytechnical University, Shenzhen 518057, China. Email: xlqian@nwpu.edu.cn. 
% }
% \thanks{Zixuan Wang, Haoran Sun, Jiaming Lu, Zhongling Huang and Dingwen Zhang are with the Brain and Artificial Intelligence Laboratory, School of Automation, Northwestern Polytechnical University, Xi'an 710021, China. Email: wzixuan@mail.nwpu.edu.cn, sunhaoran8136@mail.nwpu.edu.cn, jmlu@mail.nwpu.edu.cn, huangzhong-ling@nwpu.edu.cn, zhangdingwen2006yyy@gmail.com.}
% \thanks{Junwei Han is with the Brain and Artificial Intelligence Laboratory, School of Automation, Northwestern Polytechnical University, Xi'an 710021, China, and also with the School of Artificial Intelligence, Chongqing University of Posts and Telecommunications, Chongqing 400065, China. Email: junweihan2010@gmail.com.}
}



\maketitle

\begin{abstract}
Infrared small target detection remains challenging due to limited feature representation and severe background interference, resulting in sub-optimal performance. While recent CLIP-inspired methods attempt to leverage textual guidance for detection, they are hindered by inaccurate text descriptions and reliance on manual annotations. To overcome these limitations, we propose DGSPNet, an end-to-end language prompt-driven framework. Our approach integrates dual-granularity semantic prompts: coarse-grained textual priors (\textit{e.g.}, ``infrared image'', ``small target'') and fine-grained personalized semantic descriptions derived through visual-to-textual mapping within the image space. This design not only facilitates learning fine-grained semantic information but also can inherently leverage language prompts during inference without relying on any annotation requirements. By fully leveraging the precision and conciseness of text descriptions, we further introduce a text-guided channel attention (TGCA) mechanism and text-guided spatial attention (TGSA) mechanism that enhance the model’s sensitivity to potential targets across both low- and high-level feature spaces. Extensive experiments on IRSTD-1K, NUDT-SIRST, and NUAA-SIRST datasets demonstrate that our method significantly improves detection accuracy and achieves state-of-the-art performance on these benchmarks.
\end{abstract}

\begin{IEEEkeywords}
Infrared small target detection, dual-granularity semantic prompt, prompt-driven semantic guidance
\end{IEEEkeywords}

\section{Introduction}
\IEEEPARstart{I}{nfrared} small target detection (IRSTD) aims to accurately localize tiny objects in infrared imagery captured by aerial or satellite platforms \cite{chen2013local,li2025enhanced}. 
These targets usually lack clear structure or texture because of thermal diffusion, which makes them hard to distinguish from complex backgrounds \cite{li2024model}.
The problem becomes even more difficult due to sensor noise, atmospheric interference, and rapidly changing scenes.
Despite these challenges, IRSTD is essential for many mission-critical tasks, such as early-warning military systems, infrastructure monitoring and disaster response. As a result, there has been growing interest in developing more reliable and robust detection methods \cite{lin2024learning,zhang2022isnet}.

Existing IRSTD methods \cite{yuan2024sctransnet,fan2024diffusion,li2026langsurf,zhang2025mamba,tang2025r2ploc} improve detection by enhancing target saliency using technologies like spatial–channel transformers or detail-preserving convolutional networks.
These approaches strengthen feature representation and highlight potential target regions, but they mainly depend on visual information extracted from images.
However, solely relying on the image modality becomes a limitation in infrared scenarios, since targets often have very low signal-to-noise ratios, weak contrast, and almost no structural details \cite{sun2023infrared,zhang2025unsupervised,gao2026cosurfgs}.
Consequently, they may perform poorly in cluttered or dynamic environments, which shows the need for more adaptive and context-aware solutions.

\IEEEpubidadjcol
With the emergence of multi-modal visual–language models (VLMs)~\cite{huang2025text}, recent studies have explored leveraging text descriptions as semantic guidance to improve infrared small target detection. Textual descriptions have exceptional advantages of being highly abstract and semantically informative, with even a short sentence able to convey information such as the overall status and target distribution in an infrared image. For example, SAIST~\cite{zhang2025saist} uses a large language model to generate text descriptions, which are then used to guide a Segment Anything–based decoder for more accurate target detection. Text-IRSTD~\cite{huang2025text} introduces fuzzy semantic text prompts and a progressive cross-modal decoder to enhance feature fusion between infrared images and textual semantics.
Although these approaches have achieved advanced progress, we notice several challenges in following this pipeline.
\textbf{(1) Annotation overhead.} Creating accurate text descriptions (for example, ``a small aerial target against a sky–ground background'') requires expert knowledge and significant manual effort. Automatically generating such descriptions with VLMs also demands a high computational cost, which limits scalability for large-scale IRSTD applications.
\textbf{(2) Description mismatch.} Text annotations, whether annotated by humans or generated by models, often fail to accurately capture characteristics required for IRSTD. On the one hand, different annotators or tools may emphasize different semantic aspects, leading to inconsistent guidance. On the other hand, the gold standard for which descriptions are instructive in distinguishing targets from complex backgrounds is ambiguous.

To embrace the advantages of multi-modal integration and address the aforementioned challenges, we propose an end-to-end framework for language-guided infrared small target detection. 
Template-based prompts, such as those used in CLIP~\cite{radford2021learning} or Text-IRSTD~\cite{huang2025text} (\textit{e.g.}, ``an infrared image with a small target''), are easy to obtain and require no annotation overhead, but they lack precise, instance-specific descriptions tailored to infrared imagery. In contrast, generating rich, image-specific textual descriptions via large language models (\textit{e.g.}, ChatGPT) provides comprehensive semantic information, yet incurs significant computational cost. To reconcile these trade-offs, we first introduce the \textit{Dual-Granularity Semantic Prompts} module. Here, ``dual-granularity'' refers to the coarse-grained textual priors and the fine-grained semantic descriptions. 
Specifically, coarse-grained textual priors (\textit{e.g.}, ``infrared image'' or ``small target'') supply general domain-aware context, while the fine-grained semantic descriptions are generated through a visual-to-text mapping within each image to capture target-specific cues. This dual design enriches the feature space with complementary semantic information, enabling more discriminative feature representations for small targets and improving localization accuracy.

Furthermore, we present the \textit{Prompt-driven Semantic Guidance} module, which fully and naturally leverages the dual-granularity semantic prompts to steer the infrared small target detection process. This guidance is realized through the text-guided channel and spatial attention mechanisms. Concretely, text-guided channel attention is integrated into the encoder to refine the extracted features according to the semantic cues, while text-guided spatial attention is applied in the decoder to enhance the accuracy of the predicted target masks. 
These semantic guidance modules direct the model to focus on semantically relevant regions while suppressing background interference, thereby improving target saliency and enhancing localization precision.
\textbf{Our contributions can be summarized as follows:}
\begin{itemize}
\item We introduce a novel dual-granularity semantic prompting strategy that combines coarse-grained textual priors with fine-grained, instance-specific semantic descriptions generated via visual-to-text mapping, enriching feature representations for small infrared targets.
\item We propose a Text-Guided Channel and Spatial Attention mechanism that incorporates refined semantic cues into the model, effectively guiding the network to focus on semantically relevant regions while suppressing background interference, thereby improving target saliency and localization accuracy.
\item Extensive experiments on three IRSTD benchmarks demonstrate that our proposed method consistently outperforms state-of-the-art methods, achieving superior detection performance in complex infrared scenarios.
\end{itemize}

\section{Related Work}
\subsection{Infrared Small Target Detection}
Early infrared small target detection (IRSTD) methods were heavily reliant on handcrafted priors designed to enhance faint targets against noisy thermal backgrounds. Techniques such as filter-based methods, including Top-Hat and Laplacian-of-Gaussian (LoG) transformations, were commonly used to highlight target features by emphasizing differences in intensity \cite{bai2010analysis, bai2010enhancement,gao2026pys}. Additionally, local contrast-based approaches that exploited regional saliency \cite{deshpande1999max,qian2026adalreid} and sparse representation models for decomposing target and background signals \cite{wu2024rpcanet,qian2026seeing} were also frequently applied. While these classical methods were computationally efficient, they proved highly sensitive to background clutter and often struggled in complex infrared scenes featuring non-uniform noise or dynamic structures, where distinguishing small targets from the background became difficult. With the advent of deep learning, convolutional neural networks (CNNs) revolutionized IRSTD by offering more robust detection. Methods like ACM, DNANet, ALCNet and YOLO-MST  \cite{dai21acm, DNANet, dai2021attentional, yue2025yolo} utilized techniques such as multiscale fusion, encoder-decoder consistency, and detail-preserving designs to improve accuracy in detecting small targets. Recent works further explore hierarchical feature fusion and multi-task learning, using ternary-tree structure with an auxiliary density map branch~\cite{wang2025hierarchical} to preserve shallow-layer target cues. However, despite their advancements, CNN-based models still largely rely on low-level texture cues and lack strong semantic reasoning, leading to poor generalization when faced with ambiguous or unseen conditions. 

To address these shortcomings, Transformer-based architectures, such as SCTransNet and DCFR-Net \cite{yuan2024sctransnet, fan2024diffusion}, were introduced. These models employ self-attention mechanisms to capture long-range dependencies and global context, which enhances feature representation, especially in cluttered scenes. Beyond static decoding, IrisNet~\cite{qian2025irisnet} introduces a meta-learned decoder that dynamically generates decoder parameters conditioned on the input image status, improving adaptability across diverse scenarios. However, even these Transformer-based models remain limited by their reliance on purely visual information, which restricts their semantic expressiveness and adaptability to diverse imaging conditions. The limitations of both classical methods and deep learning models highlight the need for more comprehensive approaches that can incorporate multi-modal or semantic information to better adapt to the complexities of infrared target detection.

\subsection{Language-guided IRSTD}
Vision–Language Models (VLMs), such as CLIP~\cite{radford2021learning} and ALIGN \cite{jia2021scaling}, have made significant strides by aligning images and text in a shared embedding space using contrastive learning. This alignment has led to remarkable performance across a wide range of tasks, including open-vocabulary detection (e.g., ViLD \cite{gu2021open}, GLIP \cite{li2022grounded}), image captioning, visual question answering, and dense prediction through multimodal frameworks like OFA \cite{wang2022ofa}, BLIP-2 \cite{li2023blip}, and UniVL \cite{zhou2021unified}. Moreover, generative models like GIT \cite{wang2022git} treat vision-language tasks as conditional generation problems, while instruction-following detectors, such as DetGPT \cite{pi2023detgpt}, combine large language models with detection heads to enable natural-language-driven vision tasks. Despite their tremendous success in the realm of RGB imagery \cite{wang2025image}, VLMs face significant challenges when adapted to infrared data, primarily due to modality gaps, such as low signal-to-noise ratios, weak target contrast, and the distinct physics behind infrared imaging. These differences make it difficult for VLMs to effectively interpret and detect targets in infrared images, limiting their performance in infrared small target detection (IRSTD). 

\added{To bridge the modality gap, recent efforts have explored SAM or LLMs for IRSTD to incorporate more realistic infrared semantics. SAM-SPL~\cite{11172325} adapts SAM's general knowledge in a consult-guide manner and generates self-derived prompts from shallow-layer features to complement low-level details. SAIST~\cite{zhang2025saist} integrates textual and visual modalities by using GPT-4V to generate visual descriptions and refining them through object-object similarity and object-scene relevance, and uses them to guide a SAM-based decoder with infrared imaging equations. These approaches demonstrate the potential of foundation models in providing semantic guidance for IRSTD.} 
Text-IRSTD~\cite{huang2025text} introduces fuzzy semantic prompts and a Progressive Cross-modal Semantic Interaction Decoder to dynamically align textual and visual features, improving target localization. 
However, these approaches still face challenges such as high annotation or computational overhead and an inability to effectively capture infrared-specific attributes, such as low-intensity thermal signals and background noise. These limitations often result in semantic mismatches between the textual descriptions and the infrared image content, which in turn degrade the overall detection performance. As a result, there remains a need for more efficient and precise methods that can bridge these gaps and fully leverage language-guided infrared detection.

\section{Methodology}

\subsection{Overview}

\begin{figure*}[t]
    \centering  
    \includegraphics[width=0.95\textwidth]{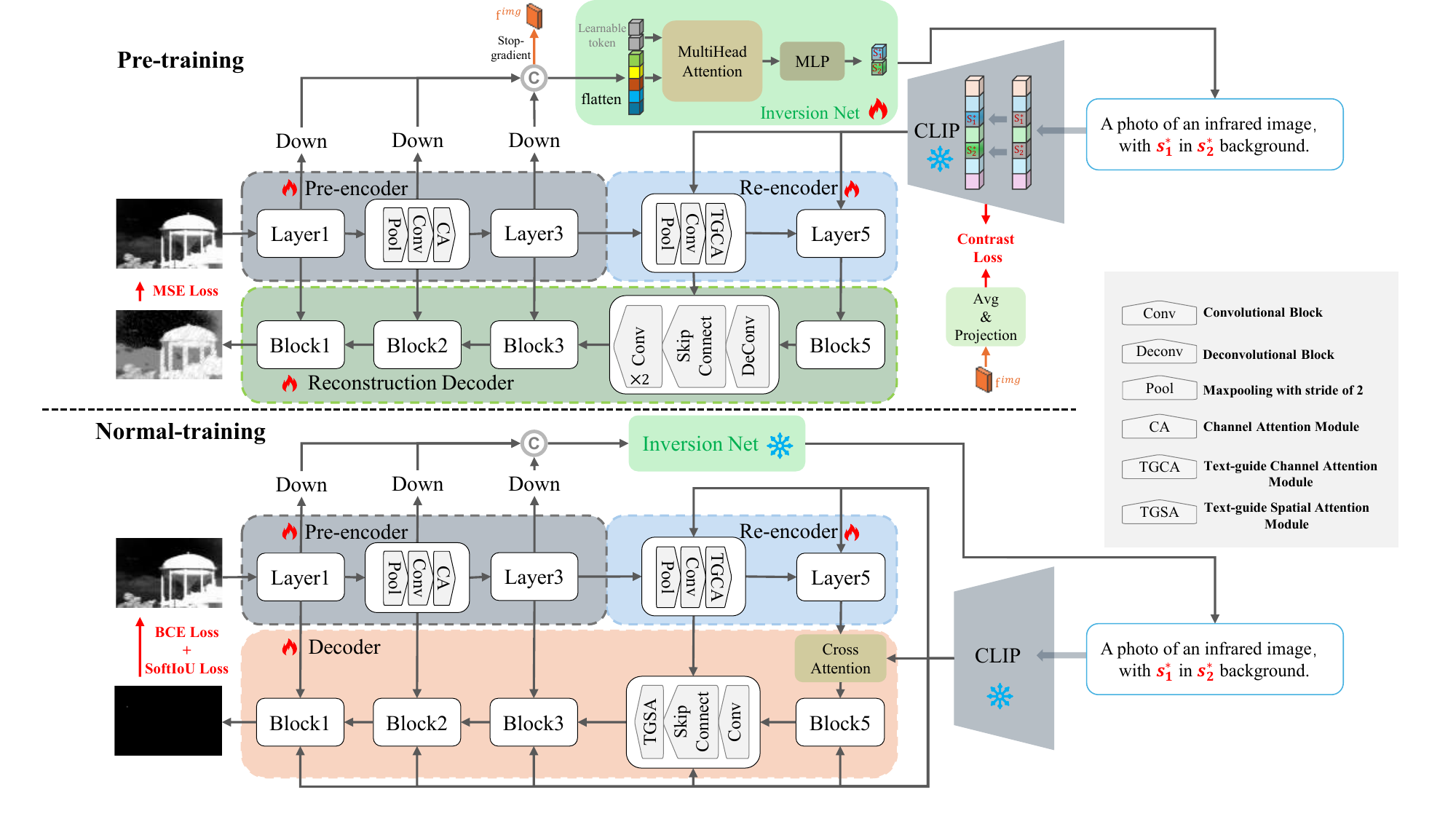}
    \vspace{-0.2in}
    \caption{Schematic of our proposed DGSPNet.
    Given an input infrared image, the pre-encoder first extracts multi-scale visual features, which are then converted into the textual space through an inversion net. The outputs, act as fine-grained implicit prompts, are integrated with a coarse-grained explicit template to form our dual-granularity semantic prompts. The semantic prompts are subsequently fed into the re-encoder and decoder to guide the learning and refinement of channel and spatial information, respectively. The illustration is organized around our two-stage training pipeline. The pre-training stage jointly trains all components using a reconstruction objective. In the normal-training stage, the inversion network is frozen, and the remaining modules are further fine-tuned.
    }
    \label{fig:overview}
    \vspace{-0.1in}
\end{figure*}

Figure~\ref{fig:overview} illustrates the schematic overview of our proposed DGSPNet, which is composed of three modules: a hierarchical image encoder, a dual-granularity semantic prompt module, and a prompt-driven semantic guidance module. Given an input infrared image, \textit{the hierarchical encoder} first extracts multi-scale visual features through a pre-encoder. \textit{The dual-granularity semantic prompt module} then projects these features into the textual space, via an inversion net, to generate fine-grained descriptions. Combined with the coarse-grained text templates, we form dual-granularity semantic prompts that capture both global priors and instance-specific semantics, and adopt a pre-trained CLIP to extract the corresponding text embeddings.
Lastly, \textit{the prompt-driven semantic guidance module} leverages the text embeddings to guide feature refinement and mask prediction. A re-encoder performs text-guided channel attention to enhance semantically relevant feature responses, while the decoder introduces text-driven spatial attention to guide the generation of the final mask, enabling accurate localization of small infrared targets.
Our DGSPNet is trained in a two-stage manner. During pre-training, the model is optimized with an image reconstruction objective to learn stable visual–textual mappings. During normal-training, the inversion net is frozen, and the remaining components are fine-tuned using detection losses. 

In the following sections, we first introduce the hierarchical image encoder to capture global contextual features in Sec.~\ref{subsec:pre-encoder}. Next, we describe the procedure of dual-granularity semantic prompts, covering both coarse-grained priors (explicit descriptions) and fine-grained semantics (implicit descriptions) in Sec.~\ref{subsec:dual-granularity}. Then, we present the prompt-driven decoding strategy, which integrates channel and spatial attention mechanisms to model interactions between image features and text embeddings in Sec.~\ref{subsec:semantic guidance}.
Finally, we detail the training and inference pipelines of DGSPNet in Sec.~\ref{subsec:training-inference}.

\begin{figure}[t]
    \centering  
    \includegraphics[width=0.95\linewidth]{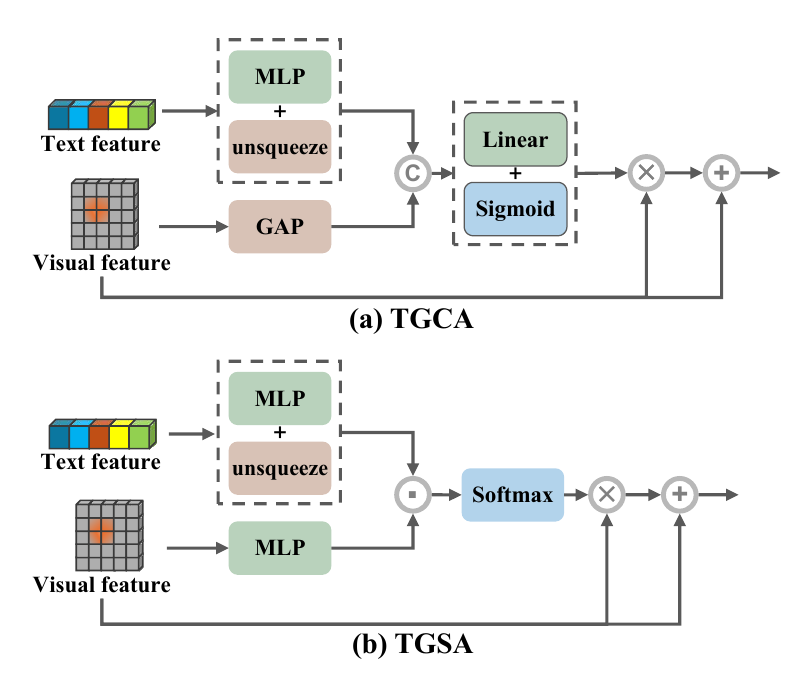}
    \vspace{-0.2in}
    \caption{Illustration of TGCA and TGSA modules. Guided by text features, these modules generate channel and spatial attention weights, which are applied via multiplication and residual connections to enhance visual features during the encoding and decoding process.} 
    \label{fig:TGCA&TGSA}
    \vspace{-0.1in}
\end{figure}

\subsection{Hierarchical Image Encoder \label{subsec:pre-encoder}}

The hierarchical encoder comprises two components of a pre-encoder and a re-encoder. The pre-encoder extracts basic multi-scale visual features, which provide the foundation for both the re-encoder and decoder to capture low-level features from the infrared images and the inversion net to generate fine-grained semantic prompts. The re-encoder then leverages dual-granularity semantic prompts to guide high-level feature refinement based on the outputs of the pre-encoder. Following the data flow, we first describe the structure of the pre-encoder in this subsection and defer the re-encoder to Sec.~\ref{subsec:semantic guidance}.

\noindent \textbf{Pre-encoder}.
Given an input infrared image $\bm{x} \in \mathbb{R}^{H \times W}$, the pre-encoder consists of three convolutional blocks that progressively extract multi-scale features $\left\{ f^{(i)}  \right\}_{i=1}^{3} \in \mathbb{R}^{h_{i} \times w_{i} \times d_{i}}$, where $h_{i}$ and $w_{i}$ denote the spatial resolution and $d_{i}$ denotes the channel dimension of the $i$-th feature map.
Except for the first block, each block performs spatial downsampling, resulting in a final feature map $f^{(3)}$ whose resolution is reduced to one-quarter of the input image.

In each block, given an input feature map $f^\text{in}$, max pooling is first applied to reduce its spatial resolution by half, followed by two stacked $3 \times 3$ convolutional layers to extract deeper representations. To enhance feature representation and capture context information contained in the image, a standard channel attention layer is appended at the end of each block to produce the refined feature map
$f^\text{out}$. This attention layer combines global average pooling (GAP) and global max pooling (GMP) to compute channel-wise attention weights, which are expected to emphasize informative channels and suppress redundant or noisy responses. The overall process can be formulated as,
\begin{equation}
\begin{aligned}
f^\text{mid} &= \gamma\left(\text{Conv}_{3\times 3}\left(\gamma(\text{Conv}_{3\times 3}(\text{MaxPool}(f^\text{in})))\right)\right) \\
f^\text{out} = &f^\text{mid} \cdot \sigma\left(\text{Conv}_{1\times 1}(\text{GMP}(f^\text{mid}) + \text{GAP}(f^\text{mid})  )\right)
\end{aligned}
\end{equation}
\noindent where $\text{Conv}_{k \times k}$ denotes a convolutional layer with a $k \times k$ kernel, and $\gamma$ and $\sigma$ represent the rectified linear unit (ReLU) and sigmoid activation functions, respectively.

\subsection{Dual-Granularity Semantic Prompts \label{subsec:dual-granularity}}

Small targets in infrared images often lack clear structural details, making it difficult to rely solely on visual cues for accurate detection, especially in complex backgrounds. Previous studies~\cite{huang2025text} have shown that incorporating information from the text-embedding space can enhance the expressiveness and semantic understanding of image features. However, generating accurate and relevant text descriptions in a cost-effective manner remains challenging. Unlike prior approaches that rely on prompt engineering or large pre-trained language models, we propose a dual-granularity semantic prompt module. 
This module, on the one hand, utilizes pre-defined text templates to provide coarse-grained general information about infrared images, and on the other hand, leverages text inversion technology to produce fine-grained attribute descriptions that are well aligned with the visual content. Together, the coarse-grained and fine-grained prompts form a complementary dual-granularity prompt. The former offers stable, general prior knowledge to guide the learning process, while the latter injects personalized, image-specific semantics to dynamically adapt the detection process. The construction of the two prompts is described in detail below.

\noindent \textbf{Coarse-grained Textual Descriptions.}
The coarse-grained textual description is designed to capture readily available prior knowledge about the input image, such as imaging modality (\textit{e.g.}, infrared) and scene type (\textit{e.g.}, sky or suburb). This information can be obtained directly from metadata or generated using predefined templates, avoiding the need for additional manual annotations. 
\added{At inference time, the scene term is selected according to the available metadata or a predefined rule, and the corresponding template is fed into the CLIP text encoder.} 
Following prior work~\cite{radford2021learning}, we design prompt templates specifically for infrared small target detection. An example prompt is:

\textit{``A photo of an infrared image, with targets in the sky/ground/ocean background.''}

These coarse-grained prompts are expressed in natural language and processed by a CLIP text encoder to produce text embeddings $f^\text{text} \in \mathbb{R}^{l \times d_{t}}$, where $l$ denotes the token sequence length and $d_{t}$ is the embedding dimension.
As global priors, coarse-grained prompts provide initial semantic context and could guide the model to focus on task-relevant regions.

\noindent \textbf{Fine-grained Semantic Tokens.} 
Coarse-grained descriptions are often too general to capture image-specific details, such as scene layout (\textit{e.g.}, the sea surface) or target category (\textit{e.g.}, a ship). Intuitively, such fine-grained information is crucial for accurately localizing infrared targets. For instance, if the model is provided with the prompt ``a ship on the sea surface'', any predicted target in the sky can be confidently regarded as a false alarm and suppressed. However, these detailed descriptions vary across images and therefore cannot be predefined using fixed templates.

To address this limitation, we introduce a set of learnable semantic tokens $\{s_{i}^{*}\}|_{i=1}^{n_t} \in \mathbb{R}^{1\times d}$ that are dynamically generated for each image using a text inversion strategy, where $n_t$ denotes the number of tokens and $d$ is the embedding dimension of the CLIP text encoder. Specifically, these tokens act as potential language representations and are learned by the inversion network from fused multi-scale features extracted from all three layers of the pre-encoder. The output tokens $\{s_{i}^{*}\}$ are thus defined as fine-grained semantic tokens. 
This enables the model to derive detailed scene semantics and local target attributes directly from visual observations.

\noindent \textbf{Dual-Granularity Semantic Prompts.}
Finally, the fine-grained semantic tokens are projected into the text embedding space and inserted into the coarse-grained textual templates, transforming them into implicit, image-aligned prompts that encode both global priors and fine-grained visual details. 
The dual-granularity semantic prompts generate text descriptions that better align with image content, without requiring any manual annotation. Meanwhile, the end-to-end training enables fine-grained semantic tokens to capture more informative and task-relevant semantics from image space, improving infrared target detection.
Taking $n_t=2$ as an example, dual-granularity semantic prompts are constructed as,

\textit{``A photo of an infrared image, with $s_{1}^{*}$ in $s_{2}^{*}$ background.''}

\noindent \textbf{Inversion Net.} We next describe the structure of the inversion net, which converts visual features into semantic tokens.
Specifically, we first take three feature maps, \textit{i.e.},  $f^{(1)}$, $f^{(2)}$, and $f^{(3)}$, from the pre-encoder and apply depthwise convolutions with a kernel size of 3 and strides of 16, 8, and 4, respectively, to downsample them. These processed features are then concatenated and passed through a $1 \times 1$ convolution to produce a fused representation $f^{\text{img}}$, which captures both global semantics and local details,
\begin{equation}
\begin{aligned}
f^\text{img} &= \text{Conv}_{1\times1} \left(\text{Concat}\left( \text{DWConv}(f^{(1)},f^{(2)},f^{(3)})\right)\right)
\end{aligned}
\end{equation}

The fused feature map $f^{\text{img}}$ is then fed into the inversion net to generate fine-grained semantic tokens, bridging low- and mid-level visual cues with textual representations.
Given $f^{\text{img}} \in \mathbb{R}^{n \times d \times h \times w}$, where $n$ is the batch size and $d$ is the channel dimension, we first reshape it into a sequence form $f^{\text{img}}_{\text{seq}} \in \mathbb{R}^{(h \times w)\times n \times d}$.
After that, we introduce a set of learnable tokens 
\added{$s^{*} \in \mathbb{R}^{n_t \times 1 \times d}$}, where $n_t$ denotes the number of fine-grained semantic tokens. These tokens are used as queries in the multi-head attention (MHA), while the keys and values are formed by combining
$s^{*}$ with the visual feature sequence $f^{\text{img}}_{\text{seq}}$.
This design allows the tokens to attend to spatially discriminative visual features while also interacting with each other to capture semantic dependencies.

Through the attention mechanism, the learnable tokens capture spatially informative and semantically meaningful patterns. Finally, these tokens are passed through a multilayer perceptron to obtain fine-grained semantic tokens that can be fed into the CLIP text encoder, completing the image-to-text inversion process. The overall procedure can be expressed as,
\begin{equation}
\begin{aligned}
s^{*} &= \text{MLP}\left(\text{MHA}\left(s^{*},(s^{*},f^{\text{img}}_\text{seq}),(s^{*},f^{\text{img}}_\text{seq})\right)\right)
\end{aligned}
\end{equation}

\subsection{Prompt-driven Semantic Guidance \label{subsec:semantic guidance}}

After obtaining the dual-granularity semantic prompts, the next question is how to effectively utilize them. Different from previous approaches that typically apply text prompts only during decoding, we propose to not only dynamically guide feature decoding in the decoder only, but also modulate feature refinement in the re-encoder. This design is motivated by the fact that high-level visual features propagated through skip connections also play a critical role in the decoding process.

\noindent \textbf{Text-Guided Channel Attention in the Re-encoder.}
The re-encoder takes as input the visual features $f^{(3)}$ from the pre-encoder together with the dual-granularity semantic embeddings 
\added{$f^{\text{text}}$}.
It consists of two successive layers that progressively produce high-level feature maps $\left\{ f^{(i)}  \right\}_{i=4}^{5} \in \mathbb{R}^{h_{i} \times w_{i} \times d_{i}}$.
Similar to the pre-encoder, each re-encoder layer is composed of a max-pooling operation followed by two convolutional layers. However, instead of using standard channel attention, we introduce a specialized Text-Guided Channel Attention (TGCA), which leverages textual semantics to strengthen channel-wise feature refinement. This design enables the model to further enhance or suppress feature responses based on the textual guidance during the encoding process. The formulation of each layer can be expressed as,
\begin{equation}
\begin{aligned}
f^\text{mid} &= \gamma\left(\text{Conv}_{3\times 3}\left(\gamma(\text{Conv}_{3\times 3}(\text{MaxPool}(f^\text{in}))\right)\right) \\
f^\text{out} &= \text{TGCA}\left(f^\text{mid},f^\text{text}\right)
\end{aligned}
\end{equation}
\noindent where TGCA computes channel attention weights for visual features under the guidance of dual-granularity semantic text embeddings.
As illustrated in Fig.~\ref{fig:TGCA&TGSA}(a), TGCA first generates text-based channel attention weights from the text embeddings using an MLP. These weights are expanded to match the spatial dimensions of the visual features. In parallel, global average pooling (GAP) is applied to the visual features to obtain image-driven channel statistics. The text-guided and image-driven attention vectors are then concatenated and fused through a linear layer to produce the final channel attention weights. After sigmoid normalization, the weights are applied to the visual features in a channel-wise manner. A residual connection is added to preserve the original information and stabilize training, yielding the final text-guided feature representation. The process of TGCA \added{can} be expressed as,
\begin{equation}
\begin{aligned}
w^\text{ch} &= \sigma \left(\text{MLP}\left(\text{Concat}\left(\text{GAP}\left(f^\text{mid}\right), \text{MLP}\left(f^\text{text}\right)\right)\right)\right) \\
f^\text{out} &= f^\text{mid}+f^\text{mid}\cdot w^\text{ch}
\end{aligned}
\end{equation}

\noindent \textbf{Text-Guided Spatial Attention in the Decoder.}
The decoder begins with a cross-attention operation that aligns the text embeddings $f^\text{text}$ with the high-level visual features $f^{(5)}$ produced by the final stage of the re-encoder. This interaction establishes a global semantic correspondence between textual prompts and image features, yielding semantically guided features for subsequent decoding. 
After this alignment, the decoder follows a five-stage cascade structure. At each stage $i$, the input decoding feature $f_{d}^{(i)} \in \mathbb{R}^{\frac{h_i}{2} \times \frac{w_i}{2} \times d}$ is first upsampled via bilinear interpolation to match the spatial resolution of the corresponding encoder feature $f^{(i)} \in \mathbb{R}^{h_i \times w_i \times d_i}$. The two features are fused through element-wise addition after projecting the encoder feature to a compatible dimension using a $1\times1$ convolution,
\begin{equation}
f_\text{fused}^{(i)} = \text{Up}(f_{d}^{(i)}) + \text{Conv}_{1\times 1}(f^{(i)}).
\end{equation}

The fused feature is subsequently refined by the proposed Text-Guided Spatial Attention (TGSA) block, which incorporates the end-of-sequence ([eos]) token from the text embedding $f^{[eos]}_\text{text} \in \mathbb{R}^{1 \times d_t}$. As illustrated in Fig.~\ref{fig:TGCA&TGSA}(b), both visual and textual features are mapped into a shared latent space of dimension $d_{c}$ through separate MLPs,
\begin{equation}
    \begin{aligned}
        {v}^{(i)} &= \text{MLP}^{i}_{\text{vis}}(f_\text{fused}^{(i)})\in\mathbb{R}^{h_{i}\times w_{i}\times d_{c}}\\
{t}^{(i)} &= \text{MLP}^{i}_{\text{txt}}({f^{[eos]}_\text{text}})\in\mathbb{R}^{1 \times d_{c}}
    \end{aligned}
\end{equation}

The text vector $t^{(i)}$ is broadcast across the spatial dimensions of the visual feature, and a dot-product operation is used to compute the semantic similarity.
A softmax normalization is thus performed to produce the spatial attention weight,
\begin{equation}
    {w}^{(i)} = \text{Softmax}\left(\frac{\langle{v}^{(i)},\,{t}^{(i)}\rangle}{\sqrt{d_c}}\right)
\in\mathbb{R}^{h_{i}\times w_{i}\times 1}
\end{equation}

The resulting attention map ${w}^{(i)}$ acts as a residual gating mechanism that selectively amplifies regions likely to contain targets while suppressing background responses,
\begin{equation}
    {f}_{d}^{(i-1)} = \text{Conv}_{3\times 3}(f_\text{fused}^{(i)} + {w}^{(i)} \cdot f_\text{fused}^{(i)})
\end{equation}

The same procedure is repeated across all decoding stages, progressively decoding the feature representation under the guidance of the semantic prompt. At the final stage, the decoder produces $f_d^{(1)}$, which is fed into a shared convolutional prediction head to generate the binary target mask.

\subsection{Optimization Objectives \label{subsec:training-inference}}

Our preliminary study finds that the inversion net is difficult to train effectively from scratch, yet it is crucial for both the dual-granularity semantic prompt and the prompt-driven semantic guidance modules. To address this issue, we adopt a two-stage training strategy, which consists of a pre-training stage for parameter initialization and a normal-training stage for task-specific optimization.

\noindent \textbf{Pre-training Stage}. 
The goal of pre-training is to obtain meaningful initial weights for the inversion net while simultaneously providing a strong initialization for the encoder. During pre-training, the architectures of the encoder and inversion net remain unchanged, while the original task-specific decoder is replaced with a reconstruction decoder composed of transposed convolution layers. The model is trained to reconstruct the input infrared image, which enables it to learn informative visual representations and reasonable initial parameters. The overall pre-training objective is defined as,
\begin{equation}
\begin{aligned}
    \mathcal{L}_{pre} = & \mathcal{L}_{\text{Contra}}\left(\text{sg}[f^\text{img}], f_\text{text}^{[eos]} \right)+ \mathcal{L}_{\text{MSE}}\left(\bm{x}, \bm{\hat{x}}\right)
\end{aligned}
\end{equation}

\noindent where $\text{sg}[\cdot]$ denotes the stop-gradient operation. The mean squared error loss $\mathcal{L}_{\text{MSE}}$ measures the reconstruction difference between the generated image and the input image, ensuring that the model preserves detailed visual information. In addition, a contrastive loss $\mathcal{L}_{\text{Contra}}$ is introduced to optimize the inversion net. Specifically, the loss is computed between the fused image feature $f^\text{img}$ and the text embedding $f_\text{text}^{[eos]}$. By maximizing their similarity, the inversion net is encouraged to learn a semantically consistent mapping from visual features to the text space.

\noindent \textbf{Normal-training Stage}. After pre-training, we discard the reconstruction decoder and replace it with a semantic-guided decoder (described in Sec.~\ref{subsec:semantic guidance}). The normal-training stage adopts the same training objective as those used in conventional infrared small target detection. Specifically, DGSPNet takes an infrared image as input and produces a predicted mask $\bm{\hat{m}}$, which is trained under the supervision of the corresponding ground-truth mask $\bm{m}$. The overall loss function is,
\begin{equation}
\begin{aligned}
    \mathcal{L}_{normal} = & \lambda_{1} \mathcal{L}_{\text{BCE}}\left(\bm{m}, \bm{\hat{m}}\right) + \lambda_{2} \mathcal{L}_{\text{SoftIoU}}\left(\bm{m}, \bm{\hat{m}}\right)
\end{aligned}
\end{equation}

\noindent where $\mathcal{L}_{\text{BCE}}$ denotes the binary cross-entropy loss, which penalizes pixel-wise prediction errors, while $\mathcal{L}_{\text{SoftIoU}}$ represents the SoftIoU loss, which directly optimizes segmentation quality by encouraging overlap between the predicted and ground-truth masks. The coefficients $\lambda_{1}$ and $\lambda_{2}$ balance the contributions of the two terms.

\section{Experiments}

\begin{table*}[t]
  \centering
  \small
    \caption{Comparison with existing IRSTD approaches on the IRSTD-1k, NUDT-SIRST, and NUAA-SIRST datasets. The evaluation metrics are IoU ($10^{-2}$), $P_d$ ($10^{-2}$), and $F_a$ ($10^{-6}$).The first and second best results are highlighted in \textbf{bold} and \underline{underlined}, respectively.    \added{The symbol ``-'' denotes that the corresponding result is not reported in the original paper.}}
    \vspace{-0.1in}
    \setlength{\tabcolsep}{3mm}{
    \begin{tabular}{lccccccccccc}
    \toprule
    \multicolumn{1}{c}{\multirow{2}{*}{\textsc{Method}}} & \multirow{2}{*}{\textsc{Backbone}} & \multirow{2}{*}{\textsc{Size}} & \multicolumn{3}{c}{\textbf{IRSTD-1K}} & \multicolumn{3}{c}{\textbf{NUDT-SIRST}} & \multicolumn{3}{c}{\textbf{NUAA-SIRST}} \\
    \cmidrule(r){4-6} \cmidrule(r){7-9} \cmidrule(r){10-12}
    & & & IoU $\uparrow$ & P$_{d} \uparrow$ & F$_{a} \downarrow$ & IoU $\uparrow$ & P$_{d} \uparrow$ & F$_{a} \downarrow$ & IoU $\uparrow$ & P$_{d} \uparrow$ & F$_{a} \downarrow$ \\
    \midrule
    IPI~\cite{gao2013infrared}  & Traditional & 256 & 27.92 & 81.37 & 16.18 & 17.76 & 74.49 & 41.23 & 25.67 & 84.63 & 16.67 \\
    PSTNN~\cite{zhang2019infrared}  & Traditional & 256 & 24.57 & 71.99 & 35.26 & 14.85 & 66.13 & 44.17 & 30.30 & 72.80 & 48.99 \\
    MSLSTIPT~\cite{9203993}  & Traditional & 256 & 11.43 & 79.03 & 1524 & 8.342 & 47.40 & 888.1 & 10.30 & 82.13 & 1131 \\
    \midrule
    RKformer~\cite{zhang2022rkformer}  & Hybird & 512 & 64.12 & 93.27 & 18.65 & 92.25 & 96.86 & 6.580 & 77.24 & \textbf{99.11} & \textbf{1.580} \\
    IAANet~\cite{wang2022interior}  & Hybird & 256 & 66.25 & 93.15 & 14.20 & 90.22 & 97.26 & 8.320 & 74.22 & 93.53 & 22.70 \\
    MTU-Net~\cite{wu2023mtu}  & Hybird & 256 & 66.11 & 93.27 & 36.80 & 74.85 & 93.97 & 46.95 & 74.78 & 93.54 & 22.36 \\
    SCTransNet~\cite{yuan2024sctransnet} & Hybird & 256 & 68.03 & 93.27 & \textbf{10.74} & 94.09 & 98.62 & 4.290 & 77.50 & 96.95 & 13.92 \\
    \midrule
    ACM~\cite{dai2021asymmetric}  & CNN & 256 & 59.23 & 93.27 & 65.28 & 61.12 & 93.12 & 55.22 & 68.93 & 91.63 & 15.23 \\
    ALCNet~\cite{dai2021attentional}  & CNN & 256 & 60.60 & 92.98 & 58.80 & 64.74 & 94.18 & 34.61 & 70.83 & 94.30 & 36.15 \\
    ISNet~\cite{zhang2022isnet}  & CNN & 256 & 61.85 & 90.24 & 31.56 & 81.24 & 97.78 & 6.340 & 70.49 & 95.06 & 67.98 \\
    FC3-Net~\cite{zhang2022exploring} & CNN & 256 & 65.07 & 91.54 & 15.55 & 78.56 & 93.86 & 23.92 & 72.44 & 98.14 & 10.85 \\
    DNA-Net~\cite{li2022dense} & CNN & 256 & 65.90 & 90.91 & \underline{12.24} & 88.19 & 98.83 & 9.000 & 75.80 & 95.82 & 8.780 \\
    UIU-Net~\cite{wu2022uiu} & CNN & 256 & 66.15 & \underline{93.98} & 22.07 & 93.48 & 98.31 & 7.790 & 76.91 & 95.82 & 14.13 \\
    ISTDU~\cite{9674870} & CNN & 256 & 66.36 & 93.60 & 53.10 & 89.55 & 97.67 & 13.44 & 75.52 & 96.58 & 14.54 \\
    RDIAN~\cite{sun2023receptive} & CNN & 256 & 56.45 & 88.55 & 26.63 & 76.28 & 95.77 & 34.56 & 68.72 & 93.54 & 43.29 \\
    AGPCNet~\cite{zhang2021agpcnet} & CNN & 256 & 66.29 & 92.83 & 13.12 & 88.87 & 97.20 & 10.02 & 75.69 & 96.48 & 14.99 \\
    Dim2Clear~\cite{10091167} & CNN & 256 & 66.34 & 93.75 & 20.93 & 81.37 & 96.23 & 9.170 & 77.29 & \underline{99.10} & \underline{6.720} \\
    MMLNet~\cite{li2025multi} & CNN & 256 & 67.21 & \textbf{94.28} & 14.00 & 81.81 & 98.43 & 11.77 & \underline{78.71} & 98.88 & 25.71 \\
    L2SKNet~\cite{wu2024saliency} & CNN & 256 & 67.81 & 90.24 & 17.46 & 93.58 & 97.57 & 5.330 & 73.43 & 98.17 &  20.82 \\
    Text-IRSTD~\cite{huang2025text} & CNN & 256 & \underline{69.57} & 92.59 & 14.97 & \underline{95.84} & \textbf{99.73} & \underline{1.032} & - & - & - \\
    \midrule
    DGSPNet (\textit{Ours})& CNN & 256 & \textbf{70.87} & 93.26 & 14.23 & \textbf{96.13} & \underline{99.15} & \textbf{0.32} & \textbf{80.32} & 97.33 & 9.26 \\
    \bottomrule
    \end{tabular}
    }
  \label{tab:comparsion_SOTA}
\end{table*}

\subsection{Experimental Setup}
\noindent \textbf{Datasets.}
We evaluate our method on three public infrared small target detection datasets, NUAA-SIRST, NUDT-SIRST, and IRSTD-1K. Following previous work~\cite{yuan2024sctransnet}, both NUAA-SIRST and NUDT-SIRST are split into training and testing sets at a ratio of 5:5.
For IRSTD-1K, we adopt an 8:2 split, using 80\% of the data for training and 20\% for testing. This configuration follows common practice in the field and ensures fair comparison with existing methods.

\noindent \textbf{Implementation Details.}
All experiments are conducted on an NVIDIA GeForce RTX 4090 GPU. 
We augment training images with normalization and random crop to a resolution of $256\times 256$.
During the pre-training stage, we use the AdamW optimizer with a learning rate of 3e-4 for the inversion net and 1e-4 for the other model parameters. Pre-training is performed for 800 epochs. After this stage, the learned weights of the encoder and inversion net are preserved for the initialization of the normal-training stage, while the reconstruction decoder is discarded. During the normal-training stage, we switch to the Adam optimizer and apply a Polynomial Decay learning rate scheduler. The initial learning rate is set to 1e-4 with a decay power of 1.2, and training proceeds for 800 epochs with a batch size of 4. Note that we do not use extra datasets for pre-training. For each evaluation dataset, its training set is used for both pre-training and normal training. 
Additionally, we adopt pre-trained CLIP-ViT-B/32 as the text encoder. 
Three complementary metrics of Intersection over Union (IoU), detection probability ($P_{d}$), and false-alarm rate ($F_{a}$) are used to evaluate models.

\subsection{Comparisons with the State-of-the-Arts}

Table~\ref{tab:comparsion_SOTA} presents quantitative comparisons between our method and several state-of-the-art approaches across three benchmark datasets. Overall, our method consistently achieves better performance, delivering higher IoU scores while maintaining comparable or lower false-alarm rates $F_{a}$. Notably, methods that obtain higher IoU than ours typically suffer from increased false alarms, whereas those with lower $F_{a}$ tend to exhibit inferior IoU, highlighting the balanced performance of our model.
More concretely, DGSPNet achieves an IoU of $70.87$ on IRSTD-1K, surpassing the second-best method Text-IRSTD~\cite{huang2025text}, which obtains $69.57$. It also reaches a high detection probability $P_{d}$ of $93.26$ while keeping the $F_{a}$ at $14.23$, demonstrating strong detection capability with controlled false positives. On NUDT-SIRST, DGSPNet delivers better performance with an IoU of 96.13, a detection probability of 99.15, and an extremely low $F_{a}$ of 0.32. These results significantly outperform competing methods and indicate highly accurate target localization together with effective background suppression. As for NUAA-SIRST, DGSPNet also achieves competitive results, obtaining an IoU of 80.32, which reflects strong detection completeness and overall accuracy.

Importantly, although DGSPNet is built on a CNN backbone, it still outperforms several methods based on hybrid CNN–Transformer architectures. For instance, on NUDT-SIRST, SCTransNet~\cite{yuan2024sctransnet} reports 94.09 IoU and 4.29 $F_{a}$, whereas DGSPNet improves the IoU to 96.13 while reducing the false-alarm rate to 0.32. This clear margin demonstrates that our proposed designs enable models to fully exploit feature extraction and semantic fusion, while further enhancing their capability through language-guided learning.

\begin{table*}[t]
  \centering
  \caption{Ablation study on the prompt-driven semantic guidance module to evaluate the necessity and synergy of sub-components (\textit{i.e.}, TGCA, Cross-Attention, TGSA). Experiments are conducted on IRSTD-1K and NUDT-SIRST datasets, with evaluation metrics including IoU ($10^{-2}$), $P_d$ ($10^{-2}$), and $F_a$ ($10^{-6}$).}
  \vspace{-0.1in}
  \small
  \setlength{\tabcolsep}{4.2mm}
  \begin{tabular}{cccccccccc}
  \toprule
  \multicolumn{4}{c}{\textsc{Prompt-driven Module Variant}} & 
  \multicolumn{3}{c}{\textbf{IRSTD-1K}} & \multicolumn{3}{c}{\textbf{NUDT-SIRST}} \\
  \cmidrule(r){1-4} \cmidrule(r){5-7} \cmidrule(l){8-10}
  \textbf{Baseline} & \textbf{TGCA} & \textbf{Cross-Attention} & \textbf{TGSA} & IoU $\uparrow$ & $P_d \uparrow$ & $F_a \downarrow$ & IoU $\uparrow$ & $P_d \uparrow$ & $F_a \downarrow$ \\
  \midrule
  \checkmark & & & & 65.31 & 91.91 & 17.72 & 94.38 & 98.15 & 8.66 \\ 
  \checkmark & \checkmark & & & 66.99 & 92.25 & 20.89 & 94.43 & 98.83 & 2.69 \\ 
  \checkmark & \checkmark & \checkmark & & 67.54 & 92.92 & 26.93 & 94.66 & 98.73 & 2.64 \\
  \checkmark & \checkmark & & \checkmark & 67.87 & 92.59 & 27.97 & 95.27 & 99.04 & 2.57 \\ 
  \checkmark & & \checkmark & \checkmark & 67.81 & 91.91 & 36.00 & 94.81 & 98.94 & 1.49 \\
  \checkmark & \checkmark & \checkmark & \checkmark & 70.87 & 93.26 & 14.23 & 96.13 & 99.15 & 0.32 \\ 
  \bottomrule
  \end{tabular}
  \label{tab:ablation}
\end{table*}

\begin{table*}[t]
  \centering
  \caption{Ablation study on text designs to evaluate the influence of the number of fine-grained semantic tokens (\#Tokens) on model performance, implemented on the IRSTD-1K dataset. Various text variants (incorporating different quantities of fine-grained tokens such as \(s_1^*\), \(s_2^*\)) are assessed, with evaluation metrics including IoU($10^{-2}$), $P_d$ ($10^{-2}$), and $F_a$ ($10^{-6}$).}
  \vspace{-0.1in}
  \small
  \setlength{\tabcolsep}{6.5mm}
  \begin{tabular}{ccccc}
  \toprule
  \textbf{\#Tokens} & \textsc{Text Variant} & IoU $\uparrow$ & $P_d \uparrow$ & $F_a \downarrow$ \\
  \midrule
  0 & A photo of an infrared image, with targets in the background. & 66.41 & 91.92 & 22.43 \\ 
  1 & A photo of an infrared image, with $s_{1}^{*}$ in the background. & 69.29 & 92.25 & 17.78 \\ 
  2 & A photo of an infrared image, with $s_{1}^{*}$ in $s_{2}^{*}$ background. & 70.87 & 93.26 & 14.23 \\ 
  3 & A photo of an infrared image, with $s_{1}^{*}$ $s_{2}^{*}$ in $s_{3}^{*}$ background. & 67.37 & 93.26 & 16.45 \\
  4 & A photo of an infrared image, with $s_{1}^{*}$ $s_{2}^{*}$ in $s_{3}^{*}$ $s_{4}^{*}$ background. & 69.22 & 91.58 & 19.91 \\ 
  \bottomrule
  \end{tabular}
  \label{tab:ablation_of_prompt}
\end{table*}

\begin{table}[t]
  \centering
  \caption{Ablation study of training strategies on IRSTD-1K. Experiments verify the effects of whether to adopt pre-training and whether to freeze the inversion net, with evaluation metrics including IoU ($10^{-2}$), $P_d$ ($10^{-2}$), and $F_a$ ($10^{-6}$).}
  \vspace{-0.1in}
  \small
  \setlength{\tabcolsep}{2mm}{
  \begin{tabular}{lccc}
  \hline
  \multicolumn{1}{c}{\textsc{Method}} & IoU $\uparrow$ & P$_d \uparrow$ & F$_a \downarrow$ \\
  \hline
  \textit{w/o} pre-training & 67.78 & 93.60 & 33.78 \\
  \textit{w/o} freezing the inversion network &68.22 & \textbf{93.93} & 37.34 \\
  Ours & \textbf{70.87} & 93.26 & \textbf{14.23} \\
  \hline
  \end{tabular}}
  \label{tab:trainging strategy}
\end{table}

\iffalse
\begin{table}[t]
  \centering
    \caption{Ablation study on downsampling strategies. The evaluation metrics are IoU ($10^{-2}$), $P_d$ ($10^{-2}$), and $F_a$ ($10^{-6}$). The best results are highlighted in \textbf{bold}.}
    \setlength{\tabcolsep}{4mm}{
    \begin{tabular}{lcccc}
    \toprule
    \multicolumn{1}{c}{\multirow{2}{*}{\textsc{DwonSample Method}}} & \multicolumn{3}{c}{\textbf{IRSTD-1K}} \\
    \cmidrule(r){2-4}
    & IoU $\uparrow$ & P$_{d} \uparrow$ & F$_{a} \downarrow$ \\
    \midrule
    Nearest & 69.54 & 92.93 & 33.42  \\
    Non-overlapping Conv. & 69.44 & \textbf{94.61} & 24.29  \\
    Depth-wise Conv. & \textbf{70.87} & 93.26 & \textbf{14.23}  \\
    \bottomrule
    \end{tabular}}
  \label{tab:downsample method}
\end{table}
\fi

\begin{table}[t]
  \centering
  \caption{\added{Ablation study on the pre-training loss weights. \(\lambda_{rec}\) and \(\lambda_{contra}\) denote the weights of the reconstruction loss and the contrastive loss, respectively. The evaluation metrics are IoU (\(10^{-2}\)), \(P_d\) (\(10^{-2}\)), and \(F_a\) (\(10^{-6}\)). The best results are highlighted in \textbf{bold}.}}
  \setlength{\tabcolsep}{4.5mm}{
  \begin{tabular}{cccccc}
    \toprule
    \(\lambda_{rec}\) & \(\lambda_{contra}\) & IoU \(\uparrow\) & \(P_d\) \(\uparrow\) & \(F_a\) \(\downarrow\) \\
    \midrule
    1 & 0.5 & 69.91 & 92.78 & 15.64 \\
    1 & 1 & \textbf{70.87} & \textbf{93.26} & \textbf{14.23} \\
    2 & 1 & 69.75 & 92.51 & 16.89 \\
    1 & 2 & 68.92 & 91.83 & 21.47 \\
    \midrule
    1 & 0 & 68.83 & 92.25 & 20.26 \\
    0 & 1 & 66.60 & 91.91 & 39.01 \\
    \bottomrule
  \end{tabular}}
  \label{tab:pretrain_loss_ablation}
\end{table}

\begin{table}[t]
  \centering
    \caption{\added{Ablation study on dimensions of the encoder. The evaluation metrics are IoU ($10^{-2}$), $P_d$ ($10^{-2}$), and $F_a$ ($10^{-6}$). ``Param.'' refers to the number of parameters (M) of the U-Net architecture. The best results are highlighted in \textbf{bold}.}}
    \setlength{\tabcolsep}{2.3mm}{
    \begin{tabular}{lcccc }
    \toprule
    \multicolumn{1}{c}{\multirow{2}{*}{\textsc{Encoder Dimension}}} & \multicolumn{4}{c}{\textbf{IRSTD-1K}} \\
    \cmidrule(r){2-5}
    & IoU $\uparrow$ & P$_{d} \uparrow$ & F$_{a} \downarrow$ & Param. \\
    \midrule
    (8,  16,  32,  64,  128)  & 70.49 & 90.57 & 23.64 & 1.16 \\
    (16, 32,  64,  128, 256)  & 70.02 & \textbf{93.60} & 17.97 & 2.23 \\
    (32, 64,  128, 256, 512)  & \textbf{70.87} & 93.26 & \textbf{14.23} & 6.39 \\
    (64, 128, 256, 512, 1024)  & 67.51 & 89.90 & 24.46 & 22.78\\
    \bottomrule
    \end{tabular}}
  \label{tab:encoder dimension}
\end{table}

\begin{table}[t]
  \centering
    \caption{Ablation study on the input of the inversion net. The evaluation metrics are IoU ($10^{-2}$), $P_d$ ($10^{-2}$), and $F_a$ ($10^{-6}$). The best results are highlighted in \textbf{bold}.}
    \setlength{\tabcolsep}{4mm}{
    \begin{tabular}{clcccc}
    \toprule
    \multirow{2}{*}{\textsc{No.}} & \multicolumn{1}{c}{\multirow{2}{*}{\textsc{Layers}}} & \multicolumn{3}{c}{\textbf{IRSTD-1K}} \\
    \cmidrule(r){3-5}
    & & IoU $\uparrow$ & P$_{d} \uparrow$ & F$_{a} \downarrow$ \\
    \midrule
    1 & (1, 2)  & 67.70 & 89.90 & 10.68 \\
    2 & (1, 2, 3)  & \textbf{70.87} & \textbf{93.26} & 14.23 \\
    3 & (1, 2, 3, 4) & 69.39 & 91.92 & 18.56 \\
    4 & (1, 2, 3, 4, 5)  & 68.63 & 90.91 & 20.53 \\
    \bottomrule
    \end{tabular}}
  \label{tab:inversion input}
\end{table}

\begin{table}[t]
  \centering
    \caption{Ablation study on loss coefficient. $\lambda_1, \lambda_2$ denote the coefficients of BCE loss and SoftIou loss, respectively. The evaluation metrics are IoU ($10^{-2}$), $P_d$ ($10^{-2}$), and $F_a$ ($10^{-6}$). The best results are highlighted in \textbf{bold}.}
    \setlength{\tabcolsep}{3mm}{
    \begin{tabular}{lccccc}
    \toprule
    \multicolumn{2}{c}{\multirow{1}{*}{\textsc{Loss coefficients Variant}}} &
    \multicolumn{3}{c}{\textbf{IRSTD-1K}}\\
    \cmidrule(l){3-5}
    $\lambda_1$ & $\lambda_2$  & IoU $\uparrow$ & P$_{d} \uparrow$ & F$_{a} \downarrow$ \\
    \midrule
    1   & 1    & \textbf{70.87} & 93.26 & \textbf{14.23} \\
    1   & 0.5  & 69.36 & 91.92 & 19.64 \\
    0.5 & 1    & 69.04 & \textbf{94.27} & 21.67 \\
    1 & 0      & 66.01 & 91.24 & 19.01 \\
    0 & 1      & 68.53 & 90.91 & 20.19 \\
    \bottomrule
    \end{tabular}}
  \label{tab:loss coefficient}
\end{table}

\iffalse
\begin{table}[t]
  \centering
    \caption{Ablation study on epochs of pretraining. The evaluation metrics are IoU ($10^{-2}$), $P_d$ ($10^{-2}$), and $F_a$ ($10^{-6}$). The best results are highlighted in \textbf{bold}.}
    \setlength{\tabcolsep}{4.5mm}{
    \begin{tabular}{lcccc}
    \toprule
    \multicolumn{1}{c}{\multirow{2}{*}{\textsc{Pretrain Epochs}}} & \multicolumn{3}{c}{\textbf{IRSTD-1K}} \\
    \cmidrule(r){2-4}
    & IoU $\uparrow$ & P$_{d} \uparrow$ & F$_{a} \downarrow$ \\
    \midrule
    400  & 66.19 & 89.90 & 18.94 \\
    800  & \textbf{70.87} & 93.26 & \textbf{14.23}  \\
    1200  & 70.24 & \textbf{93.60} & 14.99 \\
    \bottomrule
    \end{tabular}}
  \label{tab:pretrain epoch}
\end{table}
\fi

\begin{table*}[t]
\centering
\caption{FPS Comparison of Existing Infrared Target Detection Approaches on IRSTD-1k Dataset}
\label{tab:fps_analysis}
\setlength{\tabcolsep}{1.5mm}{
\begin{tabular}{ccccccccc}
\toprule
\textsc{Methods}  & ALCNet~\cite{dai2021attentional} & ISNet~\cite{zhang2022isnet} & AGPCNet~\cite{zhang2021agpcnet} & DNANet~\cite{li2022dense} & UIUNet~\cite{wu2022uiu} & MSHNet~\cite{liu2024infrared} & L2SKNet~\cite{wu2024saliency} & DGSPNet (\textit{Ours})\\
\midrule
FPS      & 97.08 & 57.80 & 41.15 & 33.22 & 31.54 & 40.81 & 86.20 & 89.9 \\
\bottomrule
\end{tabular}}
\end{table*}

\begin{table*}[t]
\centering
\caption{Further ablation study on the design of fine-grained semantic tokens. Comparisons against three categories of alternative designs: (i) visual-only baselines with standard attention and comparable model capacity, (ii) prompt tuning-like designs with learnable but image-independent prompts, and (iii) special cases using random queries, random fine-grained tokens, or fixed manual text prompts. All variants are evaluated on the IRSTD-1K dataset using IoU, \(P_d\), and \(F_a\) as metrics.}
\label{tab:gain}
\small
\setlength{\tabcolsep}{6.5mm}
\begin{tabular}{llccc}
\toprule
 \textbf{Variant} &\textbf{Ablation Study} & \textbf{IoU $\uparrow$} & \textbf{P$_{d} \uparrow$} & \textbf{F$_{a} \downarrow$} \\
\midrule
 - & Full model & 70.87 & 93.26 & 14.23 \\
\midrule
 \multirow{3}[0]{*}{\textit{w/o} text (similar capacity)}
  & Visual-only baseline & 65.31 & 91.91 & 17.72 \\
 & Visual-only attention & 65.75 & 92.26 & 15.42 \\
 & Visual-only attention \textit{w/} pre-training & 66.12 & 88.89 & 16.26 \\
\midrule
 \multirow{2}[0]{*}{prompt tuning-like design} & Learnable query in Inversion Net & 67.83 & 91.24 & 35.18 \\
 & Learnable fine-grained tokens & 69.17 & 89.89 & 35.67 \\
\midrule
\multirow{3}[0]{*}{special cases} & Random query in Inversion Net & 66.16 & 94.27 & 37.88 \\
 & Random fine-grained tokens & 67.32 & 94.27 & 30.87 \\
 & Fixed text prompts & 66.41 & 91.92 & 22.43 \\
\bottomrule
\end{tabular}
\end{table*}

\begin{figure}[t]
    \centering  
    \includegraphics[width=0.46\textwidth]{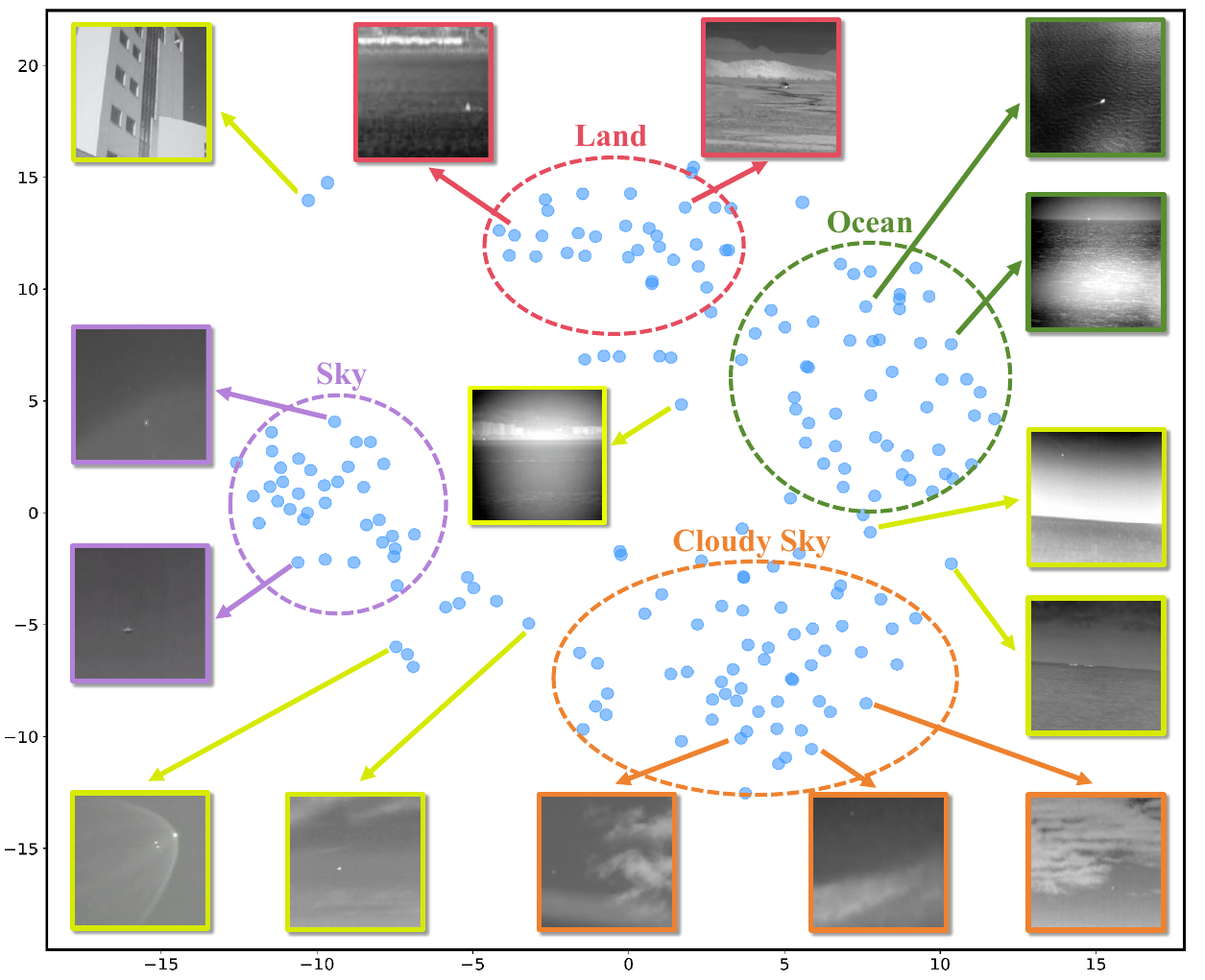}
    \vspace{-0.1in}
    \caption{t-SNE visualization of the learned fine-grained semantic token $s^{*}_{2}$ extracted from random selected samples. Images highlighted by yellow indicate samples with a mixture of background categories.}
    \label{fig:tsne}
    \vspace{-0.2in}
\end{figure}

\begin{figure*}[t]
    \centering  \includegraphics[width=0.9\textwidth]{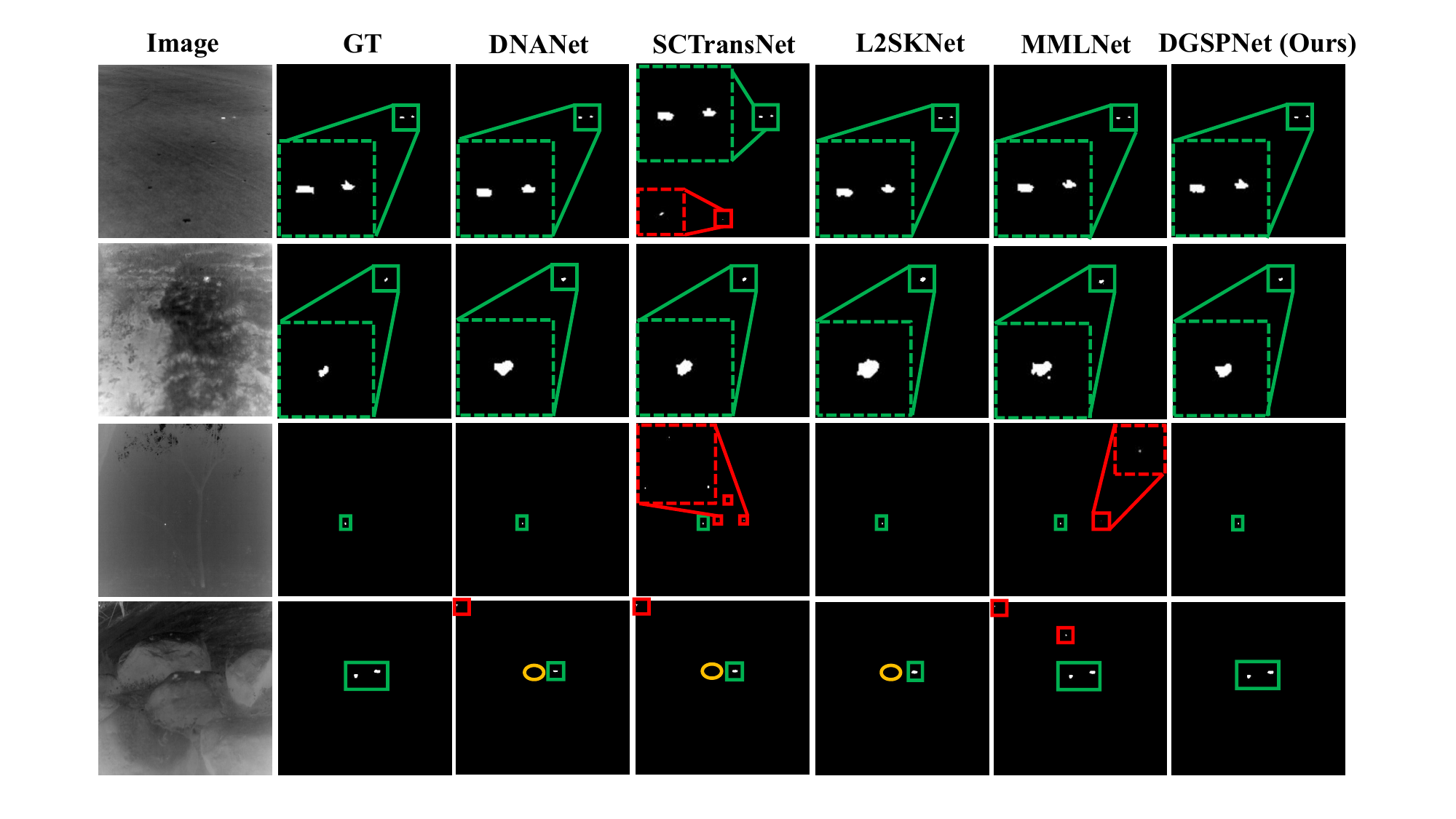}
    \vspace{-0.2in}
    \caption{Visualization of different IRSTD methods. The green boxes indicate correctly detected targets, red boxes denote false alarms, and orange circles represent missed targets. Additionally, dashed boxes show close-up views of the targets for clearer visualization.}
    \label{fig:visualization with other state-of-the-art methed}
\end{figure*}

\begin{figure*}[t]
    \centering  \includegraphics[width=0.9\textwidth]{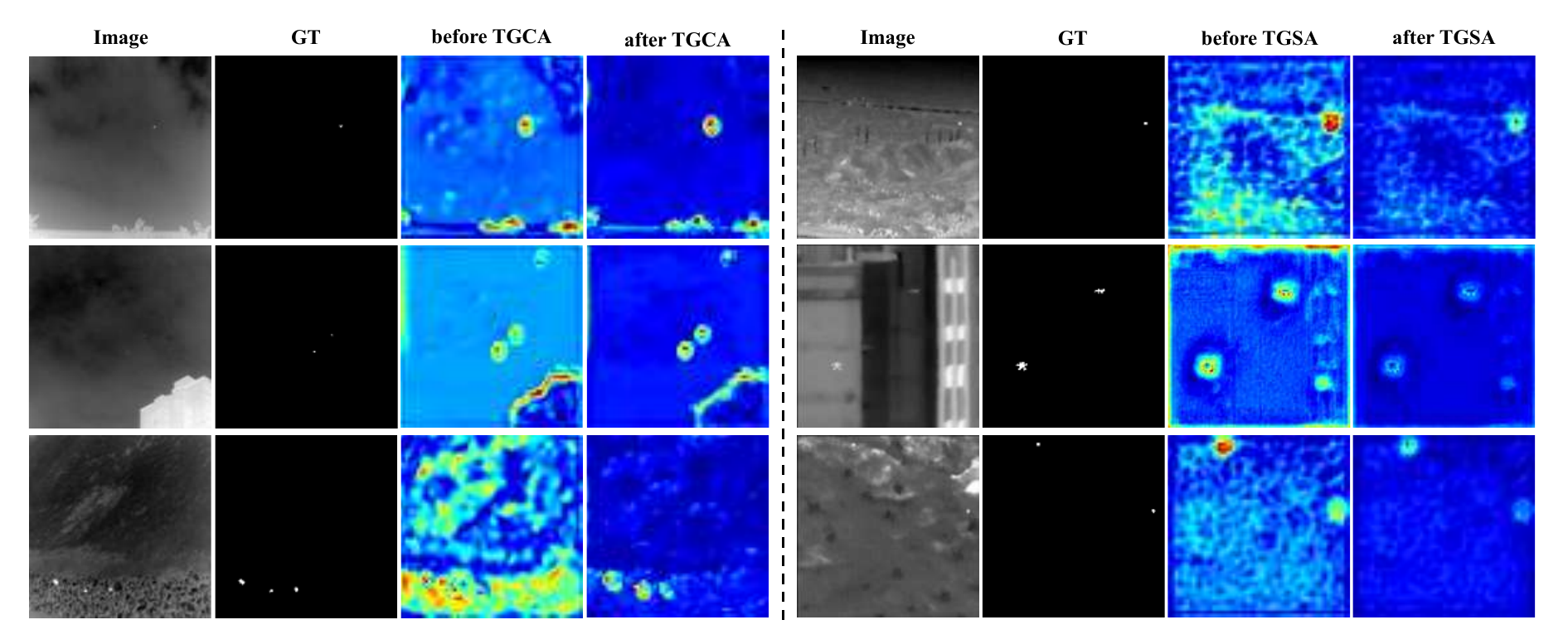}
    \vspace{-0.2in}
    \caption{\added{The left half presents the visualization of the TGCA module, while the right half shows that of the TGSA. The first column shows the original images, the second column presents the ground-truth masks, while the third and fourth columns are the feature maps before and after being fed into the TGCA and TGSA modules, respectively.}}
    \label{fig:visualization_tgca_tgsa}
\end{figure*}

\subsection{Ablation Study}

Next, we provide in-depth studies to dissect the efficacy of different designs in our framework. 

\noindent \textbf{Component Effectiveness in the Prompt-Driven Module.} 
To evaluate the contribution of each component in the prompt-driven semantic guidance module, we conduct ablation studies on text-guided channel attention (TGCA), cross-attention, and text-guided spatial attention (TGSA). By gradually adding these modules to the baseline model, we compare their effects on IoU, detection probability $F_{d}$, and false-alarm rate $F_{a}$ on the IRSTD-1K and NUDT-SIRST datasets. The results are presented in Tab.~\ref{tab:ablation}.
On IRSTD-1K, both Cross-Attention and TGSA improve IoU and $F_{d}$, indicating better target localization and more complete detections. However, these improvements come at the cost of an increased false-alarm rate. In contrast, TGCA enhances semantic alignment between text and visual features at the channel level, which effectively suppresses false alarms and reduces $F_{a}$. When all three components are combined, the model achieves the best balance among IoU, $F_{d}$, and $F_{a}$, resulting in the optimal performance on both IRSTD-1K and NUDT-SIRST. These findings demonstrate that the three modules play complementary roles and confirm the effectiveness of our design.

\noindent \textbf{Effectiveness of Dual-Granularity Semantic Prompt.} 
Table~\ref{tab:ablation_of_prompt} reports the performance of different text prompt designs for the dual-granularity semantic prompting strategy on the IRSTD-1K dataset. The results indicate the best performance is achieved when two semantic tokens are used and the prompt is formulated as ``A photo of an infrared image, with $s_{1}^{*}$ in $s_{2}^{*}$ background''.
Using only a single semantic token leads to inferior results. This suggests that a single token is usually insufficient to describe both the target and the background. In contrast, introducing too many tokens may introduce semantic ambiguity, which weakens the textual guidance.

\noindent \textbf{Designs of Training Strategy.} 
Table~\ref{tab:trainging strategy} presents an ablation study on the effects of using pre-training and freezing the inversion net during the normal-training stage. The results show that pre-training is necessary to obtain stable and reliable initialization for the inversion net. We explain that if the inversion net is trained jointly with the encoder–decoder, its supposed role might be undermined due to the presence of multiple optimization objectives. This results in a reduction of the multi-modal guiding purpose of the dual-granularity semantic prompting during the processes.

\noindent\added{\textbf{Pre-training Loss Analysis.}
To further analyze the pre-training loss design, we conduct an ablation study on the loss weights. As shown in Tab.~\ref{tab:pretrain_loss_ablation}, the reconstruction-only variant ($\lambda_{rec}=1$, $\lambda_{contra}=0$) learns useful visual representations but lacks sufficient visual-textual alignment, resulting in a moderate IoU but a relatively high false-alarm rate. The contrastive-only variant ($\lambda_{rec}=0$, $\lambda_{contra}=1$) encourages cross-modal alignment but lacks stable reconstruction supervision, which leads to a much higher false-alarm rate (39.01) and degraded detection performance. When the weights deviate from the equal setting (\textit{e.g.}, 1:0.5, 2:1, or 1:2), the performance also drops to different extents. In contrast, the equal-weight setting (1:1) achieves the best IoU (70.87), $P_d$ (93.26), and $F_a$ (14.23). This confirms that the two losses play complementary roles: the reconstruction loss stabilizes visual representation learning, while the contrastive loss improves semantic alignment between visual features and textual prompts.}

\noindent\added{\textbf{Encoder Dimension Analysis.}
The architectural details of the encoder layers are provided in Sec.~\ref{subsec:pre-encoder}. Each encoder layer starts with a $3\times 3$ convolutional block, which progressively transforms the feature representations and increases the channel dimension across layers. To determine an appropriate channel configuration for building the feature hierarchy, we conduct an ablation study on the output channel dimensions of all five encoder layers. As shown in Tab.~\ref{tab:encoder dimension}, increasing the channel dimensions generally improves model performance, as it enhances the representation capacity of the network. However, this also leads to a higher number of parameters. When the channel dimensions become too large, performance may degrade due to overfitting, especially under limited training data, while the parameter count continues to increase. Based on this trade-off, we select the channel configuration [32, 64, 128, 256, 512], which achieves a good balance between performance and false alarm rate. For practical applications with stricter constraints on model size, smaller configurations such as [16, 32, 64, 128, 256] or [8, 16, 32, 64, 128] can be used. These alternatives significantly reduce the number of parameters while maintaining competitive performance.}

\noindent \textbf{Efficacy of the Inversion Inputs.}
The inversion network requires visual features that combine fine-grained details with global contextual information. Achieving a good balance between these two aspects is the main goal of the ablation study in this section. Table~\ref{tab:inversion input} reports the results when features from different encoder layers are used as inputs to the inversion network.
When only the first layer, or the first two layers, are used, the features lack sufficient semantic depth. As a result, the inversion network cannot learn enough discriminative information, leading to degraded performance. The best performance is achieved when features from the first three layers are used, as this combination provides a good balance between detailed information and higher-level context.
However, when features from the fourth or fifth layers are further included, the performance begins to decline. This is likely because deeper layers contain more abstract and coarse information, which may dilute fine-grained details and introduce redundancy. 

\noindent \textbf{Discussions of Loss Coefficients.}
Since our method requires multiple loss functions to constrain the training process, we conduct ablation studies on the coefficients of these losses on the IRSTD-1K dataset. The text prompt is set as \textit{``A photo of an infrared image, with $s_{1}^{*}$ in $s_{2}^{*}$ background.''}. Experimental results in Tab.~\ref{tab:loss coefficient} indicate that the optimal performance on the IRSTD-1K dataset is achieved when both BCE loss and softIoU loss are assigned a coefficient of 1.

\subsection{Further Discussions on Fine-grained Semantic Tokens.} 
To fully comprehend the role played by fine-grained semantic tokens, as well as the implicit semantics they acquire from the image feature space, we provide a t-SNE visualization of $s_2^{*}$ extracted from random selected samples. 
Intuitively, $s_2^{*}$ is placed before the ``background'' noun in the dual-granularity prompt of ``A photo of an infrared image, with $s_{1}^{*}$ in $s_{2}^{*}$ background'', so it is expected to learn some of the descriptive semantics related to the background in the image. Figure~\ref{fig:tsne} shows that clustering results are indeed highly correlated with background categories. It can be seen that samples of images with sky, cloudy sky, ocean, and land backgrounds respectively have similar distances in the low-dimensional space, which fully demonstrates that $s_2^{*}$ possesses background-related semantics.

While the t‑SNE visualization qualitatively reveals that the learned fine‑grained tokens capture meaningful background semantics, it does not provide a quantitative assessment of their impact on detection accuracy, nor does it compare them with simpler, non‑adaptive prompt designs. 
We further evaluate three categories of alternative designs in Tab.~\ref{tab:gain} to dissect the source of performance gains from our fine‑grained tokens. (i) Visual‑only baselines with comparable model capacity. To isolate the contribution of textual guidance, we remove all text inputs and replace the text‑guided attention modules with standard channel/spatial attention counterparts. (ii) Prompt tuning‑style designs with learnable but image‑independent prompts. We replace our inversion‑generated tokens with two types of global learnable prompts: (a) a learnable query vector in the inversion net, and (b) learnable fine‑grained tokens that are not conditioned on visual features. (iii) Random or fixed prompt variants. To test the sensitivity to prompt content, we replace our generated tokens with random queries, random fine‑grained tokens, or fixed manual prompts (\textit{e.g.}, ``a photo of an infrared image with targets in the background''). 

Quantitatively, the visual-only baselines, even with pre-training and attention mechanisms, yield lower IoU and higher false-alarm rates, indicating that visual features alone are insufficient to resolve ambiguities in complex infrared scenes. The prompt tuning-like designs suffer from drastically increased false alarms, revealing that image-independent prompts introduce semantic noise rather than precise guidance. The random or fixed prompt variants perform even worse, with IoU ranging from 66.16 to 67.32 and $F_a$ soaring to 37.88 for random queries, suggesting that unstructured or non-adaptive prompts are unreliable for target localization. Collectively, these comparisons validate that our dual‑granularity prompting, especially the image‑adaptive fine‑grained tokens, provides discriminative semantics that effectively suppress false alarms and boost detection completeness.

\subsection{Model Complexity Analysis}

Excluding the parameters from CLIP, our U-Net model accounts for 9.26M parameters, which is a reasonable parameter count compared to other vision-language-based methods for infrared small target detection. As shown in Tab.~\ref{tab:fps_analysis}, we also compare the inference speed (FPS) of our method with several existing approaches on the IRSTD-1K dataset under an input resolution of $256 \times 256$. The results demonstrate that our method achieves superior inference speed over most competing methods.

\subsection{Qualitative Results}
Figure \ref{fig:visualization with other state-of-the-art methed} presents visualization comparisons between our method and other state-of-the-art approaches (DNANet~\cite{DNANet}, SCTransNet~\cite{yuan2024sctransnet}, L2SKNet~\cite{wu2024saliency}, MMLNet~\cite{li2025multi}). In the first and second rows, the inference results of our proposed DGSPNet are shown to be more consistent with the ground-truth masks, demonstrating the method's high localization accuracy and strong segmentation consistency. In the third and fourth rows, our method achieves zero false alarms while detecting all targets in both images, indicating that the integration of textual information can improve detection accuracy to some extent and reduce false positives.

\added{Figure~\ref{fig:visualization_tgca_tgsa} illustrates the effectiveness of the proposed TGCA and TGSA modules. The first and second columns show the original images and ground-truth masks, respectively, while the third and fourth columns display the feature maps before and after passing through the TGCA or TGSA module. A comparison between the third and fourth columns reveals that the two modules significantly suppress cluttered background information. With textual prompts, they can more precisely direct the model's attention to potential target regions. This not only validates the effectiveness of our proposed ``Prompt-driven Semantic Guidance'' strategy but also demonstrates that our training method enables the inversion net to learn correct information, thereby ensuring the validity of text inputs.}

\section{Broader Impact and Limitation}

\subsection{Broader Impact}

DGSPNet advances state-of-the-art infrared small-target detection by demonstrating that concise, automatically-generated language prompts can serve as an inexpensive supervisory signal for low-SNR imagery.  Beyond military early-warning systems, the framework is directly applicable to civilian disaster response (wildfire hotspot monitoring), maritime search-and-rescue, and infrastructure surveillance, where early detection of faint heat signatures can save both lives and resources. Because our approach requires no manual text annotations at inference time and runs efficiently at 256 × 256 resolution, it is well-suited for deployment on edge devices such as UAVs and nano-satellites.

\subsection{Limitations and Future Work}

Although our DGSPNet performs well on three IRSTD benchmarks, it still encounters a few limitations. First, our coarse-grained text description relies on a generic template of ``A photo of a ...''. More attempts at prompt engineering may help produce descriptions that lead to better IRSTD results. Second, the two-stage training pipeline inevitably increases training cost. In future work, it is expected to reduce the training overhead and improve the performance by further exploring multi-modal pre-training, alignment, and collaboration

\section{Conclusion}
We propose DGSPNet, an end-to-end framework that integrates coarse-grained textual priors with fine-grained, instance-adaptive prompts to provide explicit text guidance for infrared small-target detection. Unlike conventional approaches, DGSPNet improves detection performance by aligning hierarchical visual features with dynamically distilled semantic cues, allowing both the encoder and decoder to focus on small targets while suppressing background noise.
DGSPNet achieves state-of-the-art performance on three IRSTD datasets, demonstrating strong robustness across diverse and complex scenes. Our experiments reveal that combining fixed template prompts, which provide stable contextual priors, with learnable semantic tokens, which adapt to instance-specific characteristics, creates a complementary and effective prompting mechanism. By treating language as a compact and auxiliary signal, DGSPNet can significantly guide visual models to capture small targets in complex and noisy infrared environments.

\bibliographystyle{IEEEtran} 
\bibliography{main}

\end{document}